\documentclass{article}

    \PassOptionsToPackage{numbers, compress}{natbib}

\usepackage{multirow} 
\usepackage{rotating}
\usepackage{makecell}
\usepackage[table]{xcolor}

 \usepackage[preprint]{neurips_2026}

\usepackage[utf8]{inputenc} 
\usepackage[T1]{fontenc}    
\usepackage{hyperref}       
\usepackage{url}            
\usepackage{booktabs}       
\usepackage{amsfonts}       
\usepackage{nicefrac}       
\usepackage{microtype}      
\usepackage{xcolor}         
 \usepackage{amsmath}
\usepackage{xspace}

\usepackage{booktabs}

\newcommand{\gottalong}{Guided Observational Test-Time Adaptation\xspace}

\newcommand{\gotta}{\textsc{GOTTA}\xspace}

\usepackage{enumitem}
\usepackage{subcaption} 

\newcommand{\eg}{\textit{e.g.,}}

\DeclareMathOperator*{\argmax}{arg\,max}
\DeclareMathOperator*{\argmin}{arg\,min}
\definecolor{BestBlue}{RGB}{219,235,255}

\title{GoTTA be Diverse: Rethinking Memory Policies for Test-Time Adaptation}

\author{
Shyma Alhuwaider\thanks{Equal contribution} \quad
Yasmeen Alsaedy\footnotemark[1] \quad
Merey Ramazanova \\ 
\textbf{Silvio Giancola}  \quad  
\textbf{Bernard Ghanem} \\ \\
Center of Excellence in Generative AI, KAUST, Saudi Arabia \\
\texttt{\{shyma.alhuwaider, yasmeen.alsaedi\}@kaust.edu.sa} 
}

\begin{document}

\maketitle

\begin{abstract}
Test-time adaptation (TTA) enables a pre-trained model to adapt online to an unlabeled test stream under distribution shift.
While most TTA research focuses on the adaptation objective, practical streams also depend critically on the memory used to select which test samples drive adaptation.
Existing memory mechanisms are usually evaluated as components of specific TTA algorithms, making it difficult to isolate which memory design choices matter and when they matter.
In this work, we provide a systematic benchmark that decouples memory from the adaptation algorithm and evaluates memory policies under unified conditions across i.i.d., non-i.i.d., continual, and practical test streams.
Our study shows that effective memory management requires more than retaining recent or class-balanced samples.
In particular, intra-class diversity is a key factor for avoiding redundant buffers and maintaining representative adaptation signals under temporally correlated and label-skewed streams.
Motivated by this finding, we introduce Guided Observational Test-Time Adaptation (GOTTA), a family of diversity-aware memory policies that combine class-balanced allocation with feature-space diversity.
GOTTA memories act as drop-in replacements for existing buffers and can be paired with different TTA objectives.
Across corruption benchmarks and video-stream settings, diversity-aware memory improves adaptation most clearly under constrained memory budgets and challenging non-i.i.d. streams, while remaining competitive as memory capacity increases.
These results highlight memory management as a first-class component of robust test-time adaptation and identify diversity as a central principle for practical TTA.
\end{abstract}

\begin{figure*}[t]
  \centering
  \includegraphics[width=\textwidth]{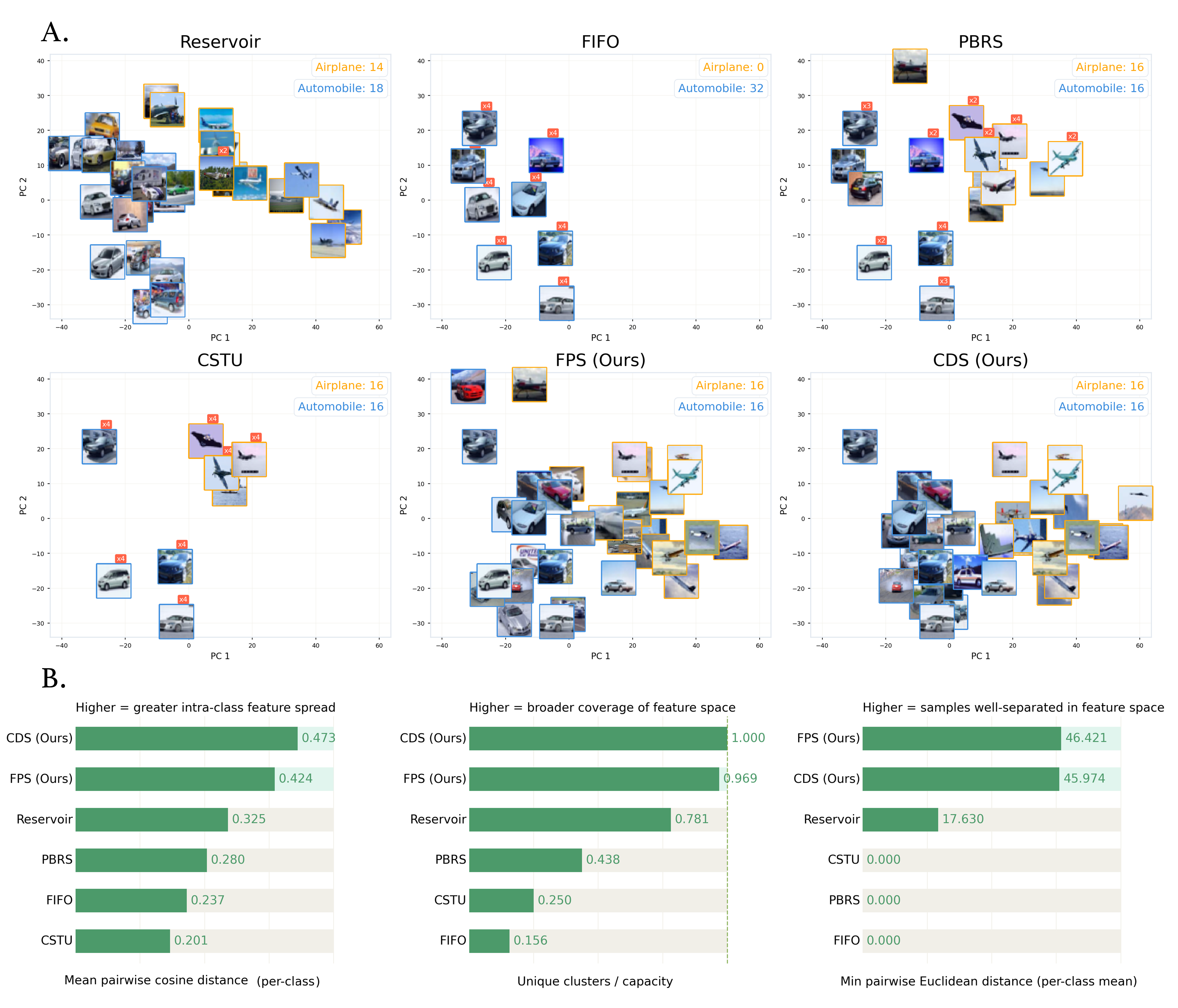}
\caption{We evaluate memory selection strategies on a two-class CIFAR-10 subset under a Dirichlet-skewed stream ($\gamma{=}0.1$) with memory capacity $M{=}32$ and $128$ streaming batches. To stress-test each strategy's ability to mitigate redundancy, every image in the dataset is duplicated four times prior to streaming, deliberately injecting near-duplicate samples into the stream.}
  \label{fig:teaser}
  \vspace{-5pt}
\end{figure*}

\section{Introduction}

Deep neural networks are typically trained under the assumption that training and test samples are independently and identically distributed~(i.i.d.).
In deployment, however, models face test distributions that shift over time due to changes in appearance, environment, acquisition conditions, or class prevalence~\citep{quinonero2008dataset}.
\textbf{Test-Time Adaptation (TTA)} addresses this problem by adapting a pre-trained, source-free model online using only the incoming unlabeled test stream, without revisiting source data~\citep{wang2018deep,zhou2022domain}.

Early TTA methods address static or mildly varying target distributions through batch normalization recalibration~\citep{A_BN,BN_Stat}, entropy minimization~\citep{tent_wang2020}, pseudo-labeling~\citep{PL}, or self-supervised objectives~\citep{ttt_sun2020}.
These methods become less reliable in practical \textbf{dynamic settings}, where test streams are non-stationary and non-i.i.d., combining gradual appearance shifts (\eg sunny $\to$ rainy) with abrupt class-prevalence changes.
In such streams, consecutive batches may be highly correlated, class-imbalanced, or visually redundant, leading to amplified prediction errors and accelerated forgetting.

Recent practical and continual TTA methods mitigate this issue with memory replay buffers.
Methods such as NOTE~\citep{note} and RoTTA~\citep{yuan2023robusttesttimeadaptationdynamic} store selected test samples to stabilize adaptation, often using class balance, confidence, uncertainty, or recency as selection criteria.
However, these memories are usually evaluated as internal components of specific TTA algorithms, making it difficult to isolate which memory design choices matter and when they matter.

\textbf{In this work, we argue that memory is a first-class, modular component of TTA.}
The memory determines which samples drive adaptation, and therefore shapes the empirical distribution seen by the model at test time.
An effective memory should not only maintain \emph{class balance}, but also preserve \emph{intra-class diversity}.
While existing memory-based TTA methods help enforce class balance, they do not explicitly prevent near-duplicate samples within each class, which can waste limited memory capacity and weaken the adaptation signal.

We therefore decouple memory from the adaptation algorithm and provide a systematic benchmark of memory policies under unified conditions.
Our analysis compares uninformed, class-guided, and diversity-aware memories across multiple TTA objectives and stream settings.
It identifies three key factors for effective memory management: \emph{class balance}, \emph{uncertainty weighting}, and, critically, \textbf{intra-class feature diversity}.
Diversity is especially important under tight memory budgets and temporally correlated streams, where each retained sample must contribute non-redundant information.

Guided by these findings, we propose \textbf{\gottalong~(\gotta)}, a family of \emph{diversity-aware memory modules} for TTA.
\gotta\ serves as a drop-in replacement for existing replay buffers and combines class-balanced allocation with intra-class geometric diversity.
It instantiates two complementary samplers: \emph{Farthest Point Sampling (FPS)} \ref{par:fps}, which promotes feature-space coverage via a nearest-neighbor diversity filter with eviction, and \emph{Cosine Diversity Sampling (CDS)} \ref{par:cds}, which promotes angular diversity within class partitions.

Figure~\ref{fig:teaser} illustrates this principle through a controlled diagnostic experiment on a two-class CIFAR-10 subset under a Dirichlet-skewed stream with duplicated samples.
Naive memories such as FIFO form dense, overlapping clusters, while class-guided memories improve balance but can still retain redundant samples.
In contrast, CDS and FPS spread retained samples more broadly across the embedding space while preserving class balance.
The corresponding diversity metrics show that these policies improve coverage and reduce near-duplicate memory entries.

These observations motivate the central design principle of \gotta: \emph{diversity and balance are not competing objectives, they can and should be enforced simultaneously}.
Our experiments show that diversity-aware memories are most beneficial under constrained memory, non-i.i.d.\ streams, and temporal correlation, while remaining competitive as memory capacity increases.
Overall, our results suggest that robust TTA requires not only effective adaptation losses, but also memory policies that control the samples used for adaptation.

\paragraph{Contributions.}
\begin{itemize}[leftmargin=*, itemsep=0pt, topsep=1pt, parsep=0pt]
    \item We decouple memory from the adaptation algorithm and evaluate memory policies under unified conditions across multiple TTA methods and stream settings.
    \item We show that memory effectiveness depends on class balance, uncertainty weighting, and intra-class feature diversity.
    \item We propose \gottalong~(\gotta), a family of diversity-aware memory modules that combine class-balanced allocation with intra-class geometric diversity for practical test-time adaptation.
\end{itemize}

\section{Related Work}

\textbf{Unsupervised domain adaptation.}
Unsupervised domain adaptation (UDA) addresses domain shift by adapting a model from a labeled source domain to an unlabeled target domain~\citep{GDCAN,wang2018deep}.
Classical UDA methods align source and target distributions through statistical distances such as Maximum Mean Discrepancy~\citep{DAN-PAMI}, adversarial domain confusion~\citep{GaninUAGLLML16,MCD}, or self-training with pseudo-labels~\citep{sepico}.
Several works further introduce class-balancing or confidence-based filtering to reduce pseudo-label bias and error accumulation~\citep{zou2018unsupervised}.
However, UDA is typically an offline setting that assumes access to source data and target samples during adaptation.
In contrast, TTA operates online, without source data, and must adapt from the incoming unlabeled test stream alone.

\textbf{Continual learning.}
Continual learning (CL) studies how models can learn from sequential tasks while mitigating catastrophic forgetting~\citep{LangeAMPJLST22}.
Major strategies include regularization-based methods that constrain important parameters~\citep{KirkpatrickPRVD16,AljundiLGB19}, parameter-isolation methods that allocate task-specific capacity, and rehearsal-based methods that replay stored examples from previous tasks~\citep{castro2018end,RebuffiKSL17}.
Our work is most related to rehearsal-based CL because it also relies on a compact memory buffer.
However, CL commonly assumes labeled task sequences, while TTA must adapt to a continuous unlabeled stream with unknown and shifting distributions.
We therefore borrow the idea of memory-based stabilization, but study it under the stricter constraints of online source-free adaptation.

\textbf{Test-time adaptation and practical streams.}
Test-time adaptation adapts a pre-trained model during inference using only unlabeled test samples~\citep{liang2025comprehensive}.
Early methods focused on static or approximately i.i.d.\ target shifts by updating batch-normalization statistics~\citep{BN,BN_Stat,A_BN}, adapting model parameters through entropy minimization~\citep{tent_wang2020}, using pseudo-labels~\citep{PL}, or relying on self-supervised objectives~\citep{ttt_sun2020}.
Other approaches adapt selected layers or the full model depending on the stability and plasticity required at test time~\citep{LiangHF20,tta_iclr1,IwasawaM21}.
While these methods are effective under standard corruption benchmarks, practical streams are often temporally correlated, non-stationary, and class-imbalanced.
Recent work therefore shifted toward continual and practical TTA settings, where distribution drift, redundancy, and forgetting become central challenges~\citep{cotta,note,yuan2023robusttesttimeadaptationdynamic}.
NOTE~\citep{note} and RoTTA~\citep{yuan2023robusttesttimeadaptationdynamic} introduced memory-based adaptation for temporally correlated streams, and ITD~\citep{alhuwaider2025advmemadversarialmemoryinitialization} further advances evaluation realism by introducing a tracklet-based video benchmark where non-i.i.d. structure is inherent rather than simulated. Our work follows this practical TTA direction, focusing specifically on the memory policy rather than proposing a new adaptation objective.

\textbf{Memory mechanisms in practical TTA.}
Memory buffers are increasingly used to stabilize adaptation under temporally correlated streams~\citep{note,yuan2023robusttesttimeadaptationdynamic,yu2024stampoutlierawaretesttimeadaptation}.
Existing policies include uninformed strategies such as FIFO and reservoir sampling~\citep{10.1145/3147.3165}, class-guided strategies that maintain pseudo-label balance~\citep{note,yuan2023robusttesttimeadaptationdynamic}, and selective filtering methods that reject unreliable or outlier samples~\citep{yu2024stampoutlierawaretesttimeadaptation}.
These designs improve stability, but they are usually introduced as components of specific TTA methods.
This makes it difficult to separate the effect of memory management from the effect of the adaptation objective.
In this paper, we decouple memory from adaptation and compare memory policies under a unified protocol, allowing us to isolate which memory properties matter across methods and streams.

\textbf{Geometric diversity and coreset selection.}
Temporally correlated streams often contain redundant samples, making memory capacity a scarce resource.
This motivates viewing memory as a compact coreset of observed test samples rather than a passive replay buffer.
Diversity-based selection has been widely studied in active learning and coreset construction, where informative subsets are chosen to cover the feature space efficiently~\citep{sener2017active,Giancola_2023_CVPR}.
Common mechanisms include farthest point sampling, which greedily maximizes the distance to the selected set~\citep{eldar1997farthest,qi2017pointnet++}, and determinantal point processes, which favor diverse subsets through repulsive sampling~\citep{taskar2013determinantal,chen2018fast}.
Inspired by these principles, we study whether feature-space diversity can improve memory management for TTA.
Unlike standard coreset selection, our setting is online, source-free, pseudo-labeled, and constrained by the evolving predictions of the adapting model.

\section{Methodology}

\subsection{Problem definition}

Let $f_{\theta_s}: \mathcal{X} \to \mathcal{Y}$ be a model pre-trained on a labeled source dataset $\mathcal{D}_s=\{(x^s,y^s)\}$ drawn from a source distribution $\mathcal{P}_s$.
In test-time adaptation, only the source model parameters $\theta_s$ are available at deployment, while the source data $\mathcal{D}_s$ is inaccessible.
The model receives a continuous unlabeled test stream $\mathcal{X}_1,\ldots,\mathcal{X}_T$ drawn from a target distribution that differs from the source distribution.
In practical settings, this target distribution is not fixed, but evolves over time as a sequence $\{\mathcal{P}_t\}_{t=1}^T$, where $\mathcal{P}_t$ may differ from $\mathcal{P}_{t-1}$ due to appearance shifts, temporal correlation, or changes in class prevalence.

At each step $t$, the model observes a batch $\mathcal{X}_t$ and updates its parameters by minimizing an unsupervised adaptation loss:
\[
    \theta_t \leftarrow \operatorname{Adapt}(\theta_{t-1}, \mathcal{X}_t; \mathcal{L}_{\mathrm{adapt}}).
\]
The updated model is then used for subsequent predictions.
Without additional control over the samples used for adaptation, this online process can be sensitive to redundant batches, class imbalance, and error accumulation in non-i.i.d.\ streams.

\subsection{Test-time adaptation with memory}

Memory-augmented TTA replaces direct adaptation on the current batch with adaptation on a curated buffer of observed test samples.
The memory $\mathcal{M}$ is a fixed-capacity buffer of size $N$.
At time $t$, it contains $K \leq N$ entries,
\[
    \mathcal{M}_t = \{m_1,\ldots,m_K\}.
\]
Each memory entry stores an observed sample and model-derived metadata:
\[
    m_i = (x_i,\hat{y}_i,u_i,z_i,a_i),
\]
where $x_i$ is the input sample, $\hat{y}_i$ is its pseudo-label, $u_i$ is its uncertainty, $z_i$ is the stored model-derived representation, and $a_i$ is its age since insertion.
For standard memory updates, $\hat{y}_i$ is obtained from the maximum softmax prediction, $u_i$ is the predictive entropy over the softmax distribution, and $z_i \in \Delta^{C-1}$ is the corresponding softmax probability vector.
These quantities are fixed at insertion time for static memory policies.
For the dynamic variant, stored samples are periodically re-forwarded through the current model to refresh their representations, pseudo-labels, and uncertainty estimates.

At each time step, memory-augmented adaptation proceeds in two stages:
\[
    \mathcal{M}_t \leftarrow \Pi(\mathcal{M}_{t-1}, \mathcal{X}_t),
\]
\[
    \theta_t \leftarrow \operatorname{Adapt}(\theta_{t-1}, \mathcal{M}_t; \mathcal{L}_{\mathrm{adapt}}),
\]
where $\Pi$ is the memory policy that determines insertion and eviction.
This formulation decouples the adaptation objective from the memory policy, allowing us to evaluate memory mechanisms independently of the underlying TTA method.

\subsection{A taxonomy of memory policies}
\label{par:taxonomy}

We organize memory policies into three families according to the information they use.
\paragraph{Uninformed memories.}
Uninformed policies ignore model predictions and representations.
FIFO keeps the most recent samples by evicting the oldest entry when memory is full.
Reservoir sampling maintains a uniformly random subset of all observed samples~\citep{10.1145/3147.3165}.
These policies are simple, but they directly inherit stream imbalance and temporal redundancy.
\paragraph{Class-guided memories.}
Class-guided policies use pseudo-labels to partition memory and encourage class balance.
PBRS applies reservoir sampling within pseudo-label partitions, preventing frequent classes from occupying the entire memory.
CSTU further combines class balancing with a timeliness and uncertainty based eviction rule~\citep{yuan2023robusttesttimeadaptationdynamic}.
These methods improve over uninformed buffers, but they do not explicitly control redundancy within each class partition.
\paragraph{Diversity-aware memories.}
We introduce \gottalong~(\gotta), a family of diversity-aware memory policies that jointly enforce class balance and intra-class diversity.
\gotta\ treats memory as a compact coreset of the observed test stream.
Rather than only balancing pseudo-label counts, it also encourages the retained samples within each class to cover the representation space.
We instantiate this principle with two policies, Cosine Diversity Sampling (CDS) and Farthest Point Sampling (FPS).

\subsection{\gottalong: diversity-aware memory for TTA}

\gotta\ is built on the principle that balance and diversity should be enforced jointly.
Each incoming sample is first assigned to a pseudo-label partition, and insertion or eviction is then decided using a combination of class allocation, uncertainty, age, and geometric redundancy.
The resulting memory remains class-balanced while avoiding near-duplicate samples within each class.



\paragraph{Shared eviction score.}

When two entries compete for the same slot, \gotta\ uses the following CSTU-style heuristic score~\citep{yuan2023robusttesttimeadaptationdynamic}:
\[
    \mathcal{H}(m_i)
    =
    \lambda_t \frac{1}{1+\exp(-a_i/N)}
    +
    \lambda_u \frac{u_i}{\log C},
    \label{eq:heuristic}
\]
where $a_i$ is the age of the entry, $u_i$ is its uncertainty, $C$ is the number of classes, and $N$ is the memory capacity.
The weights $\lambda_t$ and $\lambda_u$ control the relative importance of temporal age and uncertainty.
Larger values of $\mathcal{H}$ indicate entries that are less desirable to retain: older or more uncertain samples receive higher scores and are therefore more likely to be evicted.
When an incoming candidate competes with a stored entry, the entry with the larger score is treated as more evictable.
\paragraph{Farthest Point Sampling.}
\label{par:fps}

FPS promotes diversity by comparing each incoming sample to the closest stored sample within the same pseudo-label partition.
Although we keep the name FPS for brevity, our implementation is an efficient online nearest-neighbor diversity filter with eviction, rather than a full offline farthest point traversal.

Let $m_t=(x_t,\hat{y}_t,u_t,z_t,a_t)$ be an incoming entry.
Let $\mathcal{M}^{(\hat{y}_t)}$ denote the memory partition associated with its pseudo-label.
For each stored entry $m_i \in \mathcal{M}^{(\hat{y}_t)}$, we compute
\[
    d_i = \|z_t-z_i\|_2.
\]
If the partition has free capacity, the candidate proceeds directly to the diversity check.
If the partition is full, we identify the nearest stored entry,
\[
    m_r = \argmin_{m_i \in \mathcal{M}^{(\hat{y}_t)}} d_i.
\]

The candidate is compared only against this nearest neighbor.
If $\mathcal{H}(m_r) \geq \mathcal{H}(m_t)$, the stored neighbor is evicted and the candidate is considered for insertion.
Otherwise, the candidate is discarded.
Thus, larger values of $\mathcal{H}$ correspond to entries that are older or more uncertain, and therefore less desirable to retain.

The diversity check accepts the candidate only if
\[
    \min_i d_i > \epsilon. 
\]
where $\epsilon = 0.005$, determined empirically; we ablate its effect in the paragraph~\ref{effect_of_div}. This prevents the memory from storing samples that are too close to existing entries in the same class partition.

\paragraph{FPS with dynamic representation updates.}

FPSD extends FPS with a periodic refresh mechanism to account for the evolving feature space during adaptation. Every $\tau$ steps, all stored samples are re-encoded through the current model, updating their representations, pseudo-labels, and uncertainty estimates simultaneously. This ensures that diversity comparisons and sample selection remain consistent with the model's current understanding, rather than being anchored to an increasingly stale feature space from earlier in adaptation.

\paragraph{Cosine Diversity Sampling.}
\label{par:cds}

CDS promotes angular diversity within each pseudo-label partition.
For an incoming entry $m_t$, CDS first checks whether the corresponding partition $\mathcal{M}^{(\hat{y}_t)}$ has free capacity.
If so, the candidate is inserted directly.
If the partition is full, CDS forms the candidate set
\[
    \mathcal{C} = \mathcal{M}^{(\hat{y}_t)} \cup \{m_t\}.
\]

CDS first applies a lightweight redundancy filter.
It samples a subset $\mathcal{S} \subseteq \mathcal{M}^{(\hat{y}_t)}$ and computes cosine similarities using normalized representations,
\[
    \tilde{z}_i = \frac{z_i}{\|z_i\|_2+\varepsilon_0},
    \qquad
    \operatorname{sim}(m_t,m_i)=\tilde{z}_t^\top \tilde{z}_i.
\]
If
\[
    \max_{m_i\in\mathcal{S}} \operatorname{sim}(m_t,m_i) \geq 1-\epsilon_{\cos},
\]
the candidate is discarded as redundant, where we set $\epsilon_{\cos} = \epsilon = 0.005$. If the candidate passes the filter, CDS identifies the most redundant entry in the candidate set.
It computes the full cosine similarity matrix over $\mathcal{C}$ and finds
\[
    i^* = \argmax_i \max_{j\neq i} \tilde{z}_i^\top \tilde{z}_j.
\]
If $i^*$ corresponds to an existing memory entry, that entry is evicted and $m_t$ is inserted.
If $i^*$ corresponds to the candidate itself, the candidate is discarded.
This rule removes the most redundant element while preserving the class allocation.

\noindent\textbf{Compatibility and Modularity} The proposed memories are designed to be modular.
They do not change the adaptation loss, model architecture, or training protocol of the underlying TTA method.
Instead, they change only the empirical distribution used for test-time updates.
This makes \gotta\ compatible with different adaptation objectives and allows us to study memory management as an independent component of TTA.
\section{Experiments}

\label{sec:exp_setup}

\textbf{Datasets and streams.}
We evaluate on CIFAR-10-C
, ImageNet-C~\citep{corruptions}, and ITD \cite{alhuwaider2025advmemadversarialmemoryinitialization}, which apply 15 common corruptions across five severity levels on the test set.
Following standard TTA protocols~\citep{tent_wang2020}, we report results at severity level 5.
We consider two stream regimes: i.i.d. streams, where samples are drawn uniformly at random, and temporally correlated non-i.i.d. streams constructed using the Practical Test-Time Adaptation (PTTA) protocol~\citep{note,yuan2023robusttesttimeadaptationdynamic}, where per-batch class proportions are sampled from a Dirichlet distribution. 
Unless stated otherwise, we set the concentration parameter to $\gamma=10^{-1}$. Further implementation details for the ITD setup are provided in Appendix~\ref{app:itd_setup}

\textbf{Adaptation protocol and metrics.}
We evaluate both episodic and continual adaptation.
In the episodic setting, the model is reset to its source parameters at the beginning of each corruption.
In the continual setting, the model is updated continuously without reset across the corruptions.
We report top-1 classification accuracy averaged over the evaluated test streams and then across corruption types when multiple corruptions are used.
We fix the random seed within each stream setting to ensure controlled comparisons.


\textbf{TTA baselines and memory modules.}
We evaluate memory policies across a set of representative TTA methods: TENT~\citep{tent_wang2020}, SHOT~\citep{LiangHF20}, PL~\citep{PL}, CoTTA~\citep{cotta}, NOTE~\citep{note}, RoTTA~\citep{yuan2023robusttesttimeadaptationdynamic}, EATA/ETA~\citep{EATA}, SAR~\citep{sar}, and Energy~\citep{yuan2024teatesttimeenergyadaptation}.
We also report the \textbf{Source} model, which evaluates the pre-trained network directly on the test stream without adaptation.
For each adaptation method, we use the hyperparameters reported in the corresponding original work.
For memory, we compare policies from the three families defined in Section~\ref{par:taxonomy}.
Uninformed memories include FIFO and Reservoir~\citep{10.1145/3147.3165}.
Class-guided memories include PBRS~\citep{note} and CSTU~\citep{yuan2023robusttesttimeadaptationdynamic}.
Our \textbf{\gotta} memories include FPS, FPSD, and CDS, which combine class-balanced allocation with intra-class diversity. We also consider a no-memory baseline, which processes the stream as-is and is grouped with uninformed memories.

\textbf{Models.}
We adopt standard backbones for each benchmark: WRN-28-10-BN~\citep{wildresnet} for CIFAR-10-C, ResNet-50~\citep{resnet} for ImageNet-C, and ResNet-18 for ITD, where the backbone is pretrained on the ITD training split following~\cite{alhuwaider2025advmemadversarialmemoryinitialization}. All experiments are conducted on a single NVIDIA V100 or A100 GPU. Additional results with ViT\cite{DosovitskiyB0WZ21} on ImageNet-C are provided in Appendix~\ref{subsec:vit_imagenet_backbone}.

\subsection{Main Results}
\textbf{Memory-constrained continual adaptation.}
Table~\ref{tab:memory_constrained_all_m32} reports continual adaptation results under temporally correlated PTTA streams~\cite{note,yuan2023robusttesttimeadaptationdynamic} with a memory budget $M=32$. 
Guided observational memories provide strong gains for continual methods such as RoTTA and NOTE, while also substantially improving Norm despite its simpler normalization-based design. 
This suggests that memory-induced batch diversity can stabilize adaptation even when the adaptation rule itself is not explicitly made for non-i.i.d. streams.

\begin{table}[h]
\caption[ Memory-constrained comparison on \textbf{CIFAR-10-C} in continual temporal streams 
    with $M=32$ for all memory-based methods.
    Best results per method are 
    shown in \textbf{bold}, second-best \underline{underlined}, 
    and the best overall in \colorbox{BestBlue}{blue}.]{ Memory-constrained comparison on \textbf{CIFAR-10-C} in continual temporal streams 
    with $M=32$ for all memory-based methods.
    Best results per method are 
    shown in \textbf{bold}, second-best \underline{underlined}, 
    and the best overall in \colorbox{BestBlue}{blue}. We mark --- for no-memory baseline. \footnotemark } 
\label{tab:memory_constrained_all_m32}
\centering

\setlength{\tabcolsep}{4pt}
\renewcommand{\arraystretch}{1.1}
\resizebox{\textwidth}{!}{%
\begin{tabular}{llccccccccccc}
\toprule
Policy & Memory & Source & Norm & RoTTA & NOTE & SAR & EATA & CoTTA & TENT & PL & SHOT & Energy \\
\midrule
\multirow{3}{*}{Uninformed} & --- & 56.49 & 24.75 & 50.14 & 22.78 & \textbf{24.86} & \textbf{24.71} & 22.42 & 14.80 & 13.21 & \textbf{16.75} & \textbf{12.37} \\
 & FIFO & -- & \underline{72.63} & 68.85 & \underline{70.97} & 23.11 & 22.90 & 20.81 & 12.42 & 14.43 & 12.54 & 10.36 \\
 & Reservoir & -- & 50.50 & 40.33 & 54.20 & \underline{23.13} & 22.95 & 21.76 & 12.41 & 10.24 & 12.61 & 10.74 \\
\midrule
\multirow{2}{*}{Class-guided} & PBRS & -- & 72.50 & \underline{69.66} & 70.26 & 23.01 & 22.95 & 21.60 & \textbf{15.70} & \underline{15.44} & 14.04 & 11.30 \\
 & CSTU & -- & 71.49 & 69.23 & \textbf{72.75} & 22.96 & 22.95 & 22.11 & 15.04 & 11.18 & 13.95 & 10.87 \\
\midrule
\multirow{3}{*}{\shortstack[l]{Guided\\observational}} & CDS & -- & 70.26 & 66.47 & 60.47 & 22.96 & 22.94 & 22.50 & 13.88 & 13.16 & 14.84 & 11.10 \\
 & FPSD & -- & 69.83 & 66.30 & 65.18 & 22.96 & \underline{22.96} & \textbf{22.58} & \underline{15.61} & \textbf{15.55} & \underline{15.12} & \underline{11.34} \\
 & FPS  & -- & \cellcolor{BestBlue}\textbf{76.11} & \textbf{73.90} & 69.34 & 22.96 & 22.96 & \underline{22.56} & 13.16 & 12.10 & 13.47 & 11.16 \\
\bottomrule
\end{tabular}%
}
\end{table}
\footnotetext{Formatting for other tables follows this convention}

When paired with FPS, Norm achieves the best overall result (\textbf{76.11}); RoTTA also benefits substantially, reaching \textbf{73.90} with the same memory policy. 
Both results outperform uninformed memories such as FIFO ($72.63$) and class-guided memories such as CSTU ($72.75$). 
The gains are especially pronounced in continual temporal streams, where retaining diverse and representative samples is critical. 
Overall, these results show that diversity-aware memory is particularly beneficial under tight capacity and temporal correlation.


\begin{table*}[t]
\centering
\definecolor{BestBlue}{RGB}{219,235,255}
\setlength{\tabcolsep}{3.2pt}
\renewcommand{\arraystretch}{1.08}
\caption{
Memory-capacity scaling on \textbf{CIFAR-10-C} under temporal continual test-time adaptation. Rows report individual memory mechanisms grouped by memory-policy family, while columns report NORM, and RoTTA across memory budgets $M \in \{8,16,32,64\}$.
}
\label{tab:memory_size_scaling_temporal_pivoted}
\resizebox{\textwidth}{!}{%
\begin{tabular}{llcccccccc}
\toprule
Policy & Memory & \multicolumn{4}{c}{NORM} & \multicolumn{4}{c}{RoTTA} \\
\cmidrule(lr){3-6} \cmidrule(lr){7-10}
 &  & $M=8$ & $M=16$ & $M=32$ & $M=64$ & $M=8$ & $M=16$ & $M=32$ & $M=64$ \\
\midrule
 \multirow{2}{*}{Uninformed } & --- & 24.75 & 24.75 & 24.75 & 24.75 & \textbf{50.15} & 50.14 & 50.14 & 50.14 \\
 & Reservoir & 30.55 & 37.71 & 50.50 & 64.89 & 17.49 & 27.96 & 40.33 & 57.72 \\
\midrule
\multirow{2}{*}{Class-guided} & PBRS & 62.89 & \textbf{67.80} & 72.50 & \cellcolor{BestBlue}\textbf{76.58} & 37.67 & 56.29 & 69.66 & \cellcolor{BestBlue}\textbf{75.67} \\
 & CSTU & 51.76 & 62.61 & 71.49 & 75.46 & 32.58 & 54.18 & 69.23 & 75.21 \\
\midrule
\multirow{3}{*}{\shortstack[l]{Guided\\observational}} & CDS & 51.76 & 67.42 & 70.26 & 67.87 & 31.20 & 53.15 & 66.47 & 66.45 \\
 & FPSD & \textbf{63.97} & 69.77 & 69.83 & 74.13 & 39.30 & \textbf{59.84} & 66.30 & 71.68 \\
 & FPS & 52.44 & 67.13 & \textbf{76.11} & 76.49 & 34.85 & 55.65 & \textbf{73.90} & 75.46 \\
\bottomrule
\end{tabular}%
}
\end{table*}

\label{app:memory_budget}

\textbf{Effect of memory capacity.} Table~\ref{tab:memory_imagenet_itd_episodic_horizontal} extends our evaluation to ImageNet-C~\citep{corruptions} and ITD video clips~\citep{alhuwaider2025advmemadversarialmemoryinitialization}.
Following~\citet{yuan2023robusttesttimeadaptationdynamic}, we simulate non-i.i.d.\ test streams using a symmetric Dirichlet distribution with concentration parameter $\gamma$, where smaller values induce stronger temporal class skew.
ImageNet-C is particularly challenging because it spans $1{,}000$ classes while the memory bank contains only $M=64$ samples, leaving substantially less than one slot per class on average.
In this severely under-provisioned setting, strict class balance alone is difficult to maintain, making per-instance selection especially important.

CDS and FPS are designed for this regime: rather than passively filling class slots by recency or uncertainty alone, they explicitly favor non-redundant samples within the available memory.
This behavior is reflected in the ImageNet-C results.
CDS achieves the best RoTTA accuracy at both $\gamma=10^{-1}$ ($38.65$) and $\gamma=10^{-4}$ ($32.76$), while FPS performs best with NOTE.

On ITD video clips, non-i.i.d.\ video adaptation remains challenging.
Nevertheless, FPS performs best among the memory policies when paired with RoTTA.
This suggests that an appropriate memory policy can improve adaptation when combined with a method designed for non-i.i.d.\ streams, although the benefit depends on the dataset and adaptation objective.

\textbf{Implications of memory design.}
Our results highlight the important role of memory selection in test-time adaptation. \gotta~ methods consistently outperform classical strategies in memory-constrained regimes, particularly under continual and temporally correlated streams. As memory capacity increases, the performance gap between methods narrows, indicating that larger buffers can partially compensate for weaker selection strategies.

\begin{table}[h]
\centering
\definecolor{BestBlue}{RGB}{219,235,255}
\setlength{\tabcolsep}{3.2pt}
\renewcommand{\arraystretch}{1.08}
\caption{
Memory-equipped episodic test-time adaptation on ImageNet-C and ITD. 
ImageNet-C experiments use a ResNet-50 backbone and report episodic temporal adaptation with $M=64$ under two sampler settings, $\gamma=10^{-1}$ and $\gamma=10^{-4}$. 
ITD experiments use a ResNet-18 backbone and report episodic video adaptation under sampler settings $\gamma=10^{-1}$. 
.
 }

\label{tab:memory_imagenet_itd_episodic_horizontal}
\resizebox{\textwidth}{!}{%
\begin{tabular}{llccc|ccc|ccc}
\toprule
& & \multicolumn{3}{c|}{\textbf{ImageNet-C}, $\gamma=10^{-1}$} 
& \multicolumn{3}{c|}{\textbf{ImageNet-C}, $\gamma=10^{-4}$}
& \multicolumn{3}{c}{\textbf{ITD}, $\gamma=10^{-1}$} \\
\cmidrule(lr){3-5}
\cmidrule(lr){6-8}
\cmidrule(lr){9-11}
Policy & Memory 
& Norm & RoTTA & NOTE
& Norm & RoTTA & NOTE
& Norm & RoTTA & NOTE \\
\midrule

   \multirow{2}{*}{Uninformed }  &---
    & 26.37 & 30.78 & 11.88
    & 5.88 & 20.50 & 2.07
    &  2.22 & 5.08 & \underline{0.23} \\

    &Reservoir
    & \textbf{32.17} & 31.74 & 10.25
    & \textbf{31.59} & 27.20 & 10.79
    & 
    \underline{8.27} & 0.25 & 0.13\\
\midrule

\multirow{2}{*}{Class-guided} 
    & PBRS
    & \underline{32.08} & 29.78 & 13.64
    & \underline{31.47} & 20.91 & 8.99
    & 
    \textbf{8.87} & 0.63 & \textbf{18.11}\\

    & CSTU
    & 32.02 & \cellcolor{BestBlue}\textbf{38.65} & \underline{16.16}
    & 30.60 & \underline{32.75} & \underline{11.76}
    &  5.07 & \underline{24.17} & 0.14 
    \\

\midrule

\multirow{2}{*}{\shortstack[l]{Guided-obs.}} 
    & CDS
    & 32.02 & \cellcolor{BestBlue}\textbf{38.65} & \underline{16.16}
    & 30.50 & \cellcolor{BestBlue}\textbf{32.76} & \underline{11.76}
    &  4.59 & 18.83 & 0.14\\

    & FPS
    & 31.14 & \underline{37.78} & \textbf{18.38}
    & 29.35 & 32.48 & \textbf{15.40}
    & 4.19 & \cellcolor{BestBlue}\textbf{34.15} & 0.14
    \\

\bottomrule
\end{tabular}%
}
\end{table}

\textbf{Scaling to ImageNet-C and video streams.} Table~\ref{tab:memory_imagenet_itd_episodic_horizontal} extends our evaluation to ImageNet-C~\cite{corruptions} and ITD video clips~\cite{alhuwaider2025advmemadversarialmemoryinitialization}. Following~\cite{yuan2023robusttesttimeadaptationdynamic}, we simulate non-i.i.d.\ test streams via a symmetric Dirichlet distribution with concentration parameter $\gamma$, where smaller values induce more temporally correlated, class-skewed streams. ImageNet-C spans $1{,}000$ classes while the memory bank holds only $M = 64$ samples, meaning each class is allocated less than one slot on average. Despite this severe under-provisioning, CDS and FPS maintain competitive performance, demonstrating that per-instance diversity is a more robust and scalable selection criterion than class balance. A detailed analysis of memory budget effects is provided in Appendix~\ref{app:memory_budget}. CDS and FPS are precisely designed for this: rather than passively filling class slots by recency or uncertainty alone, they actively enforce. Ensuring that the limited slots per class capture maximally distinct representations. This pays off clearly in the results: CDS achieves the best RoTTA accuracy at both $\gamma=10^{-1}$ ($38.65$) and $\gamma=10^{-4}$ ($32.76$), while FPS leads on NOTE ($18.38$ and $15.40$), with gains most pronounced in the high-correlation regime where the memory-free baseline collapses to $2.07$. On the ITD video stream, non-i.i.d.\ video adaptation remains challenging. 
Still, FPS performs best among the memory policies in this setting. This suggests that the right memory policy, when paired with an adaptation method designed for non-i.i.d.\ streams, can lead to clear improvements. 
In particular, FPS performs best on the ITD video stream when combined with RoTTA.
    \begin{figure}[t]
    \centering



    \begin{minipage}{0.94\textwidth}
        \centering
        \includegraphics[width=\linewidth]{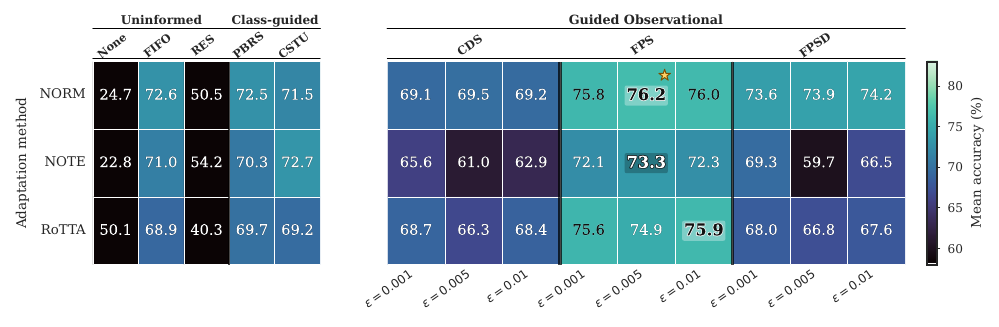}
        \label{fig:epsilon-diversity-ablation-continual}
    \end{minipage}

    \vspace{-1mm}

\caption{Memory mechanism comparison on \textbf{CIFAR-10-C} under continual temporal TTA streams with memory budget $M=32$. We compare three memory guidance classes: uninformed (None, FIFO, Reservoir), class-guided (PBRS, CSTU), and guided observational (CDS, FPS, FPSD). For guided observational methods, we ablate over diversity thresholds $\epsilon$, which control similarity between samples (Euclidean distance for FPS and cosine similarity for CDS), where larger $\epsilon$ enforces stronger diversity.
 The best result overall across all methods and strategies is indicated with a $\star$.}
    \label{fig:epsilon-diversity-ablation}
\end{figure}

\textbf{Effect of diversity threshold $\epsilon$.}
\label{effect_of_div}
We ablate the impact of the diversity threshold $\epsilon$ used in \gotta~ memory selection (CDS, FPS, FPSD), which controls the minimum dissimilarity required between stored samples. We implement $\epsilon$  as a threshold on pairwise distances (Euclidean for FPS, cosine-based for CDS), such that larger values enforce stronger diversity by preventing similar samples from being retained in memory.
Figure~\ref{fig:epsilon-diversity-ablation} shows that guided observational methods consistently outperform classical baselines across episodic and continual temporal streams under a constrained memory budget ($M=32$), with FPS achieving the strongest performance. 
This trend persists when the memory budget is increased to $M=64$ (see Appendix Figure~\ref{fig:epsilon-diversity-ablation-64}). We observe that $\epsilon = 0.005$ consistently provides the best or near-best performance across all settings (continual vs. episodic, $M=32$ vs. $M=64$) and across adaptation methods. Smaller values (e.g., $\epsilon = 0.001$) allow redundant samples to accumulate, while larger values (e.g., $\epsilon = 0.01$) can overly restrict memory updates and reduce coverage of the data distribution.
These results indicate that $\epsilon = 0.005$ strikes a favorable balance between diversity and representativeness, and we adopt it as the default setting in all experiments.
\section{Discussion and Conclusion}

We revisited the role of memory in test-time adaptation (TTA), showing that memory should be treated as an actively curated coreset rather than a passive buffer. Classical strategies based on recency or class balance can retain redundant or unrepresentative samples under realistic, temporally correlated streams. In contrast, \gottalong~(\gotta) combines class balance with intra-class diversity through FPS and CDS policies, improving adaptation across memory budgets, stream regimes, and TTA methods. Our results establish memory selection, rather than memory size alone, as a key design principle for robust and practical TTA under real-world distribution shifts.

\textbf{Limitations.}
\gotta relies on feature-space distances, whose reliability depends on representation quality during adaptation, and introduces additional curation cost relative to FIFO or reservoir memories. Our evaluation focuses mainly on corruption benchmarks and controlled video streams, leaving broader validation on deployment data for future work. Finally, the interaction between memory selection and method-specific adaptation objectives remains an open direction.

\bibliographystyle{unsrtnat}
\bibliography{sec/7_ref}
\newpage
\section{Appendix}

\subsection{Test-Time Adaptation settings}
\label{sec:tta_math}

Following~\citet{yuan2023robusttesttimeadaptationdynamic}, we distinguish four TTA settings of increasing difficulty, all sharing the same protocol: a source-trained model $f_{\theta_s}$ is adapted online using only sequential unlabeled batches $\{\mathcal{X}_t\}_{t=0}^{T}$.

\begin{itemize}[leftmargin=*, itemsep=0pt, topsep=1pt, parsep=0pt]
    \item \textbf{Fully Test-Time Adaptation (TTA)}~\citep{tent_wang2020}.
    The target distribution $\mathcal{P}_t$ is stationary and batches are i.i.d.
    The goal is a one-shot adaptation to a fixed shift $\mathcal{P}_t \neq \mathcal{P}_s$.

    \item \textbf{Continual Test-Time Adaptation (CoTTA)}~\citep{cotta}.
    The target distribution drifts as a sequence $\{\mathcal{P}_t\}$, while batches remain i.i.d.\ within each $\mathcal{P}_t$.
    The challenge is tracking distributional drift without forgetting.

    \item \textbf{Non-i.i.d.\ Test-Time Adaptation}~\citep{niid_boudiaf2022parameter,note}.
    The target distribution is stationary, but class proportions within each batch are skewed by a Dirichlet distribution $\text{Dir}(\gamma)$.
    A small concentration parameter $\gamma$ produces highly imbalanced streams where certain classes dominate.

    \item \textbf{Practical Test-Time Adaptation (PTTA)}~\citep{yuan2023robusttesttimeadaptationdynamic}.
    The most realistic setting: $\mathcal{P}_t$ evolves over time \emph{and} batches are class-imbalanced via $\text{Dir}(\gamma)$, jointly capturing temporal drift and label skew.
\end{itemize}

We follow PTTA setting in this work. 

\subsection{Additional CIFAR-10-C Experiments}
\begin{table}[h]
\caption{ Comparison on \textbf{CIFAR-10-C} under relaxed memory budget $M=64$ for memory-based methods.
    in continual temporal streams  }
\label{tab:memory_no_constraint_all_m64}
\centering
\definecolor{BestBlue}{RGB}{219,235,255}
\setlength{\tabcolsep}{4pt}
\renewcommand{\arraystretch}{1.1}
\resizebox{\textwidth}{!}{%
\begin{tabular}{llccccccccccc}
\toprule
Policy & Memory & Source & Norm & RoTTA & NOTE & SAR & EATA & CoTTA & TENT & PL & SHOT & Energy \\
\midrule
\multirow{3}{*}{Uninformed} & --- & 56.49 & 24.75 & 50.14 & 22.78 & 24.86 & 24.72 & 22.42 & 16.50 & 12.48 & 16.21 & 12.06 \\
 & FIFO & -- & 76.06 & \underline{75.52} & \underline{71.90} & 24.84 & 24.71 & 23.90 & 18.96 & \textbf{24.46} & 15.95 & 12.49 \\
 & Reservoir & -- & 64.89 & 57.72 & 62.91 & \textbf{25.11} & 24.72 & 24.07 & 14.12 & 10.41 & 14.77 & 11.77 \\
\midrule
\multirow{2}{*}{Class-guided} & PBRS & -- & \cellcolor{BestBlue}\textbf{76.58} & \textbf{75.67} & 70.37 & \underline{24.94} & 24.72 & 23.96 & \underline{19.15} & 17.03 & 15.96 & 12.06 \\
 & CSTU & -- & 75.46 & 75.21 & \textbf{72.21} & 24.75 & 24.70 & \underline{24.25} & 17.64 & \underline{22.63} & 15.90 & 12.29 \\
\midrule
\multirow{3}{*}{\shortstack[l]{Guided\\observational}} & CDS & -- & 67.87 & 66.45 & 65.47 & 24.75 & 24.73 & \textbf{24.44} & 17.30 & 14.82 & \textbf{19.88} & 12.40 \\
 & FPSD & -- & 74.13 & 71.68 & 64.00 & 24.75 & \textbf{24.74} & 24.09 & \textbf{19.57} & 20.03 & \underline{18.99} & \underline{12.67} \\
 & FPS & -- & \underline{76.49} & 75.46 & 71.09 & 24.75 & \underline{24.74} & 24.01 & 18.34 & 13.80 & 15.30 & \textbf{12.97} \\
\bottomrule
\end{tabular}%
}
\end{table}

\textbf{Scaling to additional methods.}
Table~\ref{tab:memory_no_constraint_all_m64} reports results under a memory budget ($M=64$), where capacity constraints are less restrictive, across an expanded set of TTA methods.
In this setting, we observe that guided observational memory selection remains competitive with strong baseline memory strategies. In particular, FPS (Static) achieves performance comparable to the best class-guided and uninformed methods, e.g., \underline{76.49} for Norm and 75.46 for RoTTA, closely matching PBRS (76.58, 75.67) and CSTU (75.46, 75.21). 
Unlike the memory-constrained regime, where guided observational methods provide clear gains, performance differences between memory strategies diminish as capacity increases. 
Overall, these results highlight that the primary advantage of guided observational memory lies in its ability to maintain representative and diverse samples under tight memory budgets, rather than relying on increased capacity.
    \begin{figure}[h]
    \centering

    \begin{minipage}{0.94\textwidth}
        \centering
        \includegraphics[width=\linewidth]{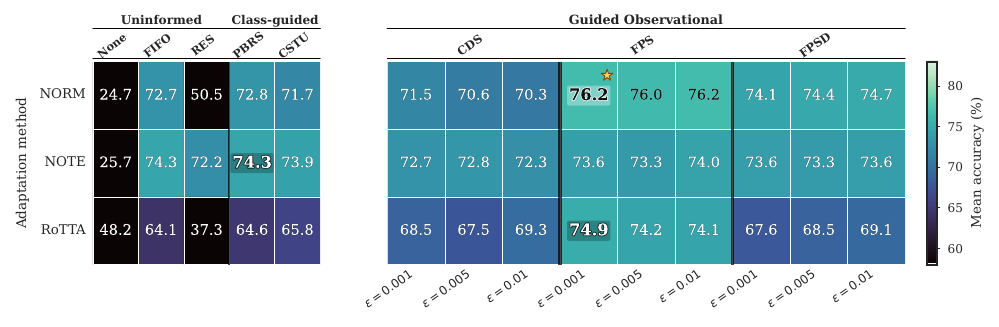}
        \label{fig:epsilon-diversity-ablation-episodic}
    \end{minipage}

    \vspace{-1mm}

\caption{Memory mechanism comparison on \textbf{CIFAR-10-C} under episodic temporal TTA streams with memory budget $M=32$. We compare three memory guidance classes: uninformed (None, FIFO, Reservoir), class-guided (PBRS, CSTU), and guided observational (CDS, FPS, FPSD). For guided observational methods, we ablate over diversity thresholds $\epsilon$, which control similarity between samples (Euclidean distance for FPS and cosine similarity for CDS), where larger $\epsilon$ enforces stronger diversity.
 The best result overall across all methods and strategies is indicated with a $\star$.}
    \label{fig:epsilon-diversity-ablation}
\end{figure}

\begin{figure}[!htbp]
    \centering

    \begin{minipage}{0.04\textwidth}
        \centering
        \rotatebox{90}{\textbf{(a) Episodic}}
    \end{minipage}%
    \begin{minipage}{0.94\textwidth}
        \centering
        \includegraphics[width=\linewidth]{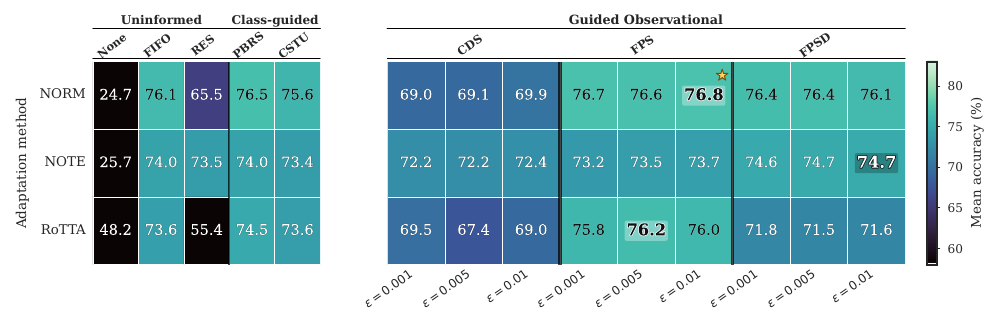}
        \label{fig:epsilon-diversity-ablation-episodic-64}
    \end{minipage}

    \vspace{-0.08cm}

    \begin{minipage}{0.04\textwidth}
        \centering
        \rotatebox{90}{\textbf{(b) Continual}}
    \end{minipage}%
    \begin{minipage}{0.94\textwidth}
        \centering
        \includegraphics[width=\linewidth]{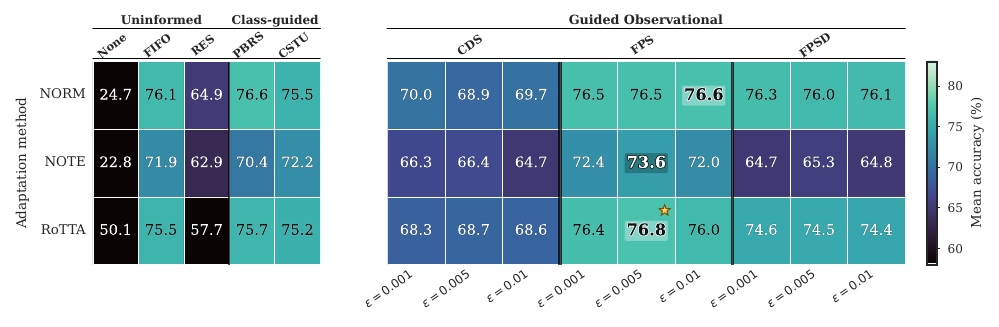}
        \label{fig:epsilon-diversity-ablation-continual-64}
    \end{minipage}

    \vspace{-1mm}

\caption{Memory mechanism comparison on \textbf{CIFAR-10-C} under episodic (a) and continual(b) temporal TTA streams with memory budget $M=64$.
The best result overall across all methods and strategies is indicated with a $\star$.}
    \label{fig:epsilon-diversity-ablation-64}
\end{figure}

\paragraph{Extended diversity threshold ablation.} Figures~\ref{fig:epsilon-diversity-ablation} and~\ref{fig:epsilon-diversity-ablation-64} extend the diversity threshold ablation to the temporal TTA stream setting.
At $M=32$ (Figure~\ref{fig:epsilon-diversity-ablation}), guided observational memories consistently outperform uninformed and class-guided baselines across all TTA methods.
FPS achieves the overall best result for NORM ($76.2\%$ at $\epsilon=0.001$) and RoTTA ($74.9\%$ at $\epsilon=0.001$), while for NOTE guided observational memories remain competitive with the best class-guided method PBRS ($74.3\%$).

At $M=64$ (Figure~\ref{fig:epsilon-diversity-ablation-64}), the same trends hold across both episodic and continual stream types.
In the episodic setting, FPS at $\epsilon=0.01$ achieves the overall best of $76.8\%$ for NORM, FPS reaches $76.2\%$ for RoTTA, and FPSD leads for NOTE at $74.7\%$.
In the continual setting, FPS at $\epsilon=0.005$ attains $76.8\%$ for RoTTA and $76.6\%$ for NORM, while FPS achieves $73.6\%$ for NOTE.
Across both memory budgets and stream types, performance remains stable over the evaluated range $\epsilon \in \{0.001, 0.005, 0.01\}$, confirming that guided observational memories are robust to the precise choice of threshold.



\label{appendix:appendix_results}

\newpage
\subsection{Effect of Memory Budget on Diversity-Aware Sample Selection}
\label{app:memory_budget}

\begin{figure*}[ht]
    \centering

    \begin{subfigure}[t]{0.5\linewidth}
        \centering
        \includegraphics[width=\linewidth]{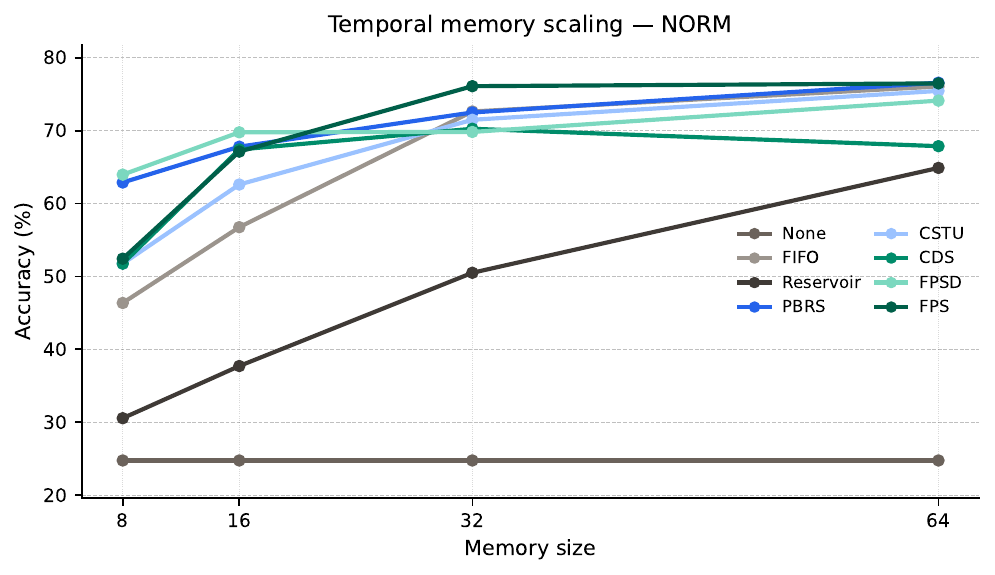}
        \caption{NORM}
        \label{fig:memory_scaling_norm}
    \end{subfigure}
    \hfill
    \begin{subfigure}[t]{0.5\linewidth}
        \centering
        \includegraphics[width=\linewidth]{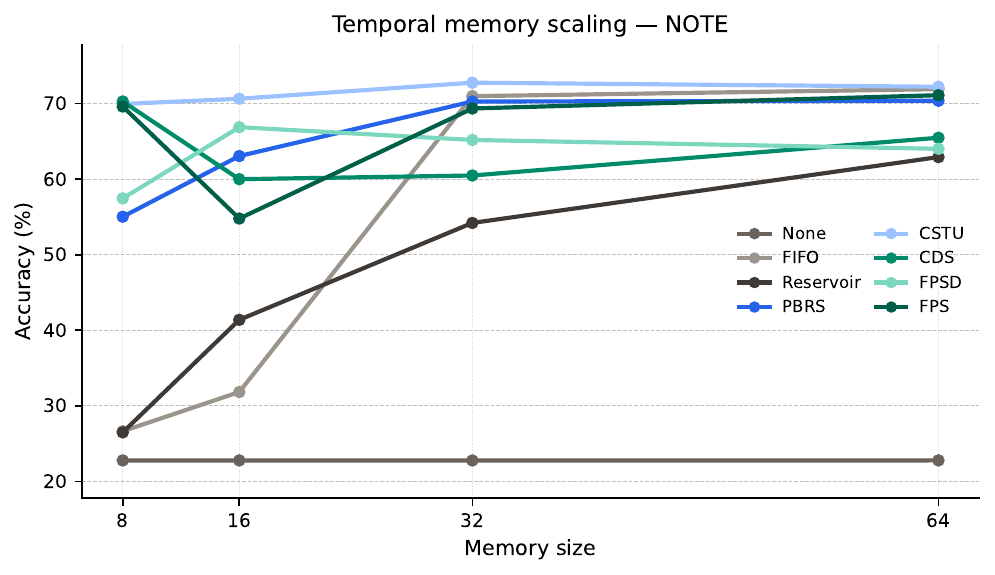}
        \caption{NOTE}
        \label{fig:memory_scaling_note}
    \end{subfigure}
    \hfill
    \begin{subfigure}[t]{0.5\linewidth}
        \centering
        \includegraphics[width=\linewidth]{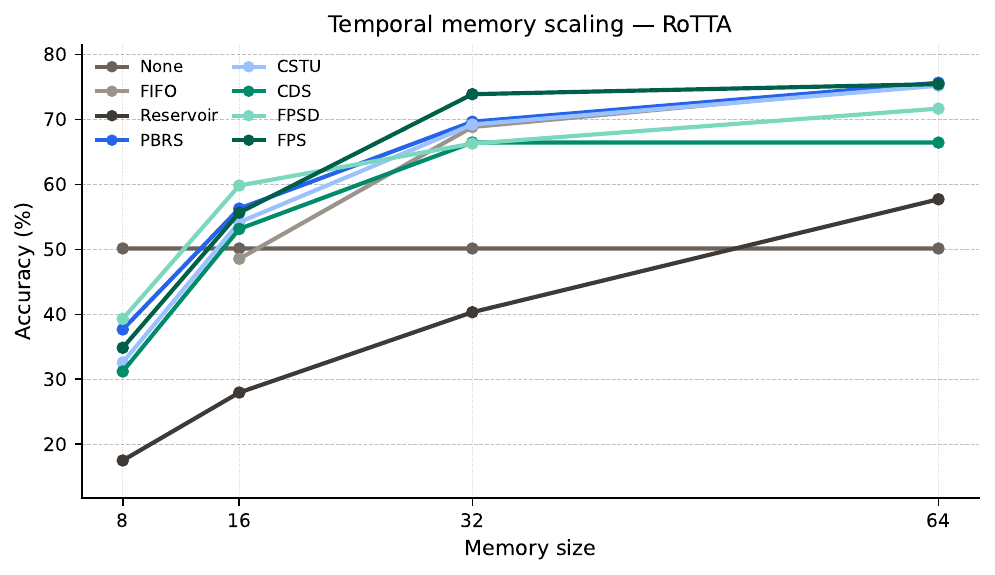}
        \caption{RoTTA}
        \label{fig:memory_scaling_rotta}
    \end{subfigure}

    \caption{
    Memory-capacity scaling under temporal continual test-time adaptation for NORM, NOTE, and RoTTA.
    Each curve shows the performance of one memory mechanism as the memory size varies from $M=8$ to $M=64$ on \textbf{CIFAR-10-C}.
    Memory mechanisms are grouped by policy family: uninformed memories, class-guided memories, and guided observational memories.
    }
    \label{fig:memory_scaling_temporal_per_method}
\end{figure*}
\paragraph{Memory capacity and diversity.}

The memory-scaling results in Figure~\ref{fig:memory_scaling_temporal_per_method} indicate that diversity-aware selection is particularly beneficial in the constrained-memory regime.
At small buffer sizes, redundant samples consume a substantial fraction of the available memory, making the choice of which samples to retain valuable.
Guided observational memories therefore provide a stronger advantage by explicitly encouraging intra-class feature diversity and maintaining a more representative buffer.
This benefit becomes less pronounced as memory capacity increases.
With larger buffers, even simpler selection strategies are more likely to retain diverse samples, reducing the relative advantage of explicit diversity enforcement.
These results suggest that the primary value of diversity-aware memory lies in improving sample efficiency: it helps the buffer remain informative when only a small number of examples can be stored.

Notably, the three TTA methods exhibit distinct sensitivity profiles to memory capacity.
For \textbf{RoTTA}, small memory budgets can be actively harmful: at $M=8$, several mechanisms perform \emph{worse} than the no-memory baseline, suggesting that a poorly populated buffer misleads the adaptation process more than it helps.
Reservoir sampling is particularly susceptible in this regime, as its i.i.d.\ assumption conflicts with the temporally correlated stream, leading to a biased and unrepresentative buffer under distribution shift.
For \textbf{NORM}, the picture is different: while any memory consistently outperforms no memory, the gap between uninformed and diversity-aware strategies remains large at small budgets, and uninformed methods require substantially larger memory capacities before approaching the performance of guided observational memories.
This underscores that for NORM, diversity-aware selection is not merely an efficiency gain but a prerequisite for competitive performance in the low-memory regime.

\subsection{ImageNet-C and ITD Evaluation Setups}
\label{app:itd_setup}

For ImageNet-C, we evaluate all methods on the standard 15 corruption types under temporal episodic test-time adaptation streams. We use a ResNet-50 backbone and report results with memory size $M=64$ under two PTTA sampler settings, $\gamma=10^{-1}$ and $\gamma=10^{-4}$.

For ITD, we evaluate episodic video adaptation using a ResNet-18 backbone pretrained on clean ITD data. To construct the video stream, we use the Gaussian noise corruption and evaluate under the PTTA setting. The stream contains 1024 video clips, with each clip consisting of 32 frames. Unless otherwise stated, all ITD experiments use memory size $M=32$ and sampler parameter $\gamma=10^{-1}$. This setup is intended to test memory-equipped adaptation under temporally coherent video streams while keeping the ITD evaluation computationally tractable.

\subsection{Backbone Generalization: ViT on ImageNet-C}
\label{subsec:vit_imagenet_backbone}

To test whether diversity-aware memory remains effective beyond convolutional backbones, we evaluate memory-equipped episodic TTA with a Vision Transformer on ImageNet-C.
This setting uses temporal streams with memory budget $M=64$ and diversity threshold $\epsilon=0.005$.
Table~\ref{tab:memory_vit_itd_episodic_horizontal} shows that diversity-aware memories remain competitive on ViT features.
For RoTTA, FPS achieves the best result, slightly improving over CSTU and CDS.
For NOTE, FPSD provides the strongest performance, suggesting that refreshing stored representations can be beneficial when the model backbone and adaptation dynamics differ from the convolutional setting.

\begin{table}[h]
\centering
\definecolor{BestBlue}{RGB}{219,235,255}
\setlength{\tabcolsep}{3.2pt}
\renewcommand{\arraystretch}{1.08}
\caption{
Memory-equipped episodic test-time adaptation on ViT ImageNet-C and ITD video clips. ViT ImageNet-C experiments use temporal streams with $M=64$ and $\epsilon=0.005$. 
}
\label{tab:memory_vit_itd_episodic_horizontal}
\begin{tabular}{llcc}
\toprule
Policy & Memory  &RoTTA & NOTE \\
\midrule
Uninformed 
 & Reservoir  & 53.67 & 47.77 \\
\midrule
Class-guided  
 & CSTU & \underline{54.02} & \underline{54.12}  \\
\midrule
\multirow{3}{*}{\shortstack[l]{Guided\\observational}} & CDS & \underline{54.02} & \underline{54.12}  \\
 & FPSD & 53.64 & \cellcolor{BestBlue}\textbf{58.38} \\
 & FPS & \textbf{54.03} & 52.84 \\
\bottomrule
\end{tabular}%
\vspace{-2mm}
\end{table}

\end{document}